\documentclass[12pt,a4]{article}
\usepackage[dvips]{color,graphicx}
\usepackage{bm,psfrag,afterpage,color,amsmath, amssymb}

\makeatletter
\def\eqnarray{\stepcounter {equation}\let \@currentlabel =\theequation
\global \@eqnswtrue
\global \@eqcnt \z@ \tabskip \@centering \let \\=\@eqncr
$$\halign to \displaywidth \bgroup \@eqnsel \hskip \@centering
$\displaystyle \tabskip \z@ {##}$&\global \@eqcnt \@ne \hfil
${\mbox{}##\mbox{}}$\hfil &\global \@eqcnt \tw@
$\displaystyle \tabskip \z@ {##}$\hfil \tabskip \@centering
&\llap {##}\tabskip \z@ \cr}
\makeatother

\begin{document}

{\baselineskip = 8mm 

\begin{center}
\textbf{\LARGE Semi-supervised logistic discrimination via labeled data and unlabeled data from different sampling distributions} \\[5mm]
\end{center}

\begin{center}
{\large Shuichi Kawano}
\end{center}

\begin{center}
\begin{minipage}{14cm}
{
\begin{center}
{\it {\footnotesize 
Department of Mathematical Sciences, Graduate School of Engineering, \\ Osaka Prefecture University, 
1-1 Gakuen-cho, Sakai, Osaka 599-8531, Japan. \\
}}

\vspace{2mm}

skawano@ms.osakafu-u.ac.jp

\end{center}

\vspace{1mm} 

{\small {\bf Abstract:} \
This article addresses the problem of classification method based on both labeled and unlabeled data, where we assume that a density function for labeled data is different from that for unlabeled data. 
We propose a semi-supervised logistic regression model for classification problem along with the technique of covariate shift adaptation. 
Unknown parameters involved in proposed models are estimated by regularization with EM algorithm. 
A crucial issue in the modeling process is the choices of tuning parameters in our semi-supervised logistic models. 
In order to select the parameters, a model selection criterion is derived from an information-theoretic approach. 
Some numerical studies show that our modeling procedure performs well in various cases. 
}

\vspace{3mm}

}
\end{minipage}
\end{center}

\baselineskip = 8mm

\vspace{3mm}


\section{Introduction}


In recent years, with the wide availability of fast and high-powered computers, high-throughput data of unexampled size and complexity have frequently been seen in the contemporary statistics and machine learning. 
Examples involve data from genomics, proteomics, natural language processing, and signal processing. 
For the huge amount of data, it is difficult to label data by a human operator, since its work requires vast times and efforts. 
Only small labeled data set may, therefore, be available, while an unlabeled data set can be more easily obtained. 
Under such a circumstance, a classification method that combines both labeled and unlabeled data, called semi-supervised learning, has received an enormous amount of attention in the late machine learning and statistical literature (see, e.g., Chapelle {\it et al.}, 2006; Liang {\it et al.}, 2007).
For overviews of semi-supervised learning methods, we refer to Zhu (2008), and references given therein.

Many classification techniques for semi-supervised learning have been proposed by various researchers, e.g., Amini and Gallinari  (2002), Basu {\it et al.}  (2004), Bennett and Demiriz  (1998), Chen and Wang  (2007), Dean {\it et al.}  (2006), Kawano and Konishi (2011), Kawano {\it et al.} (2012), Lafferty and Wasserman (2007), and Zhou {\it et al.}  (2004). 
Most of these semi-supervised methods implicitly assumes that a density function for labeled data is the same as that for unlabeled data. 
On the other hand, we, here, consider the case that the densities for labeled data and unlabeled data are different, since the densities are not always same in practical situations. 
In such a case, several semi-supervised methods have been presented, e.g., Jiang and Zhai (2007), Wu {\it et al.} (2009), and Zadrozny (2004). 
However, for these methods, there remains a problem of evaluating constructed semi-supervised models, which is a crucial issue in the model building process. 
Cross validation (CV) is often used in evaluating models constructed by semi-supervised procedures. 
An advantage of CV lies in  its independence from probabilistic assumptions.
The computational time of the procedures is, however, very large, and the high variability and tendency to undersmooth in CV are not negligible 
in the analysis of complex or high-dimensional data, since the selectors are repeatedly applied.

In this paper, we propose a logistic model for the semi-supervised classification problem by using statistical methods under covariate shift  (Shimodaira, 2000) in the case that the density function for labeled data is different from that for unlabeled data. 
The unknown parameters in the model are estimated by the regularization method  with the help of EM algorithm. 
A crucial issue in our modeling strategy is to choose values of some tuning parameters included in semi-supervised logistic models, which corresponds to evaluating models determined by our proposed procedures. 
In order to objectively select optimal values of tuning parameters, we then introduce a model selection criterion based on an information-theoretic approach (Konishi and Kitagawa, 1996) that evaluates the semi-supervised logistic models estimated by the regularization method. 
Some numerical examples demonstrate that the proposed procedure works well and performs better than competing methods.

This paper is organized as follows. 
In Section 2, we present a semi-supervised logistic model for classification problem based on covariate shift adaptation and its estimation procedure by the regularization method. 
Section 3 provides a model selection criterion derived from an information-theoretic viewpoint to select some tuning parameters in semi-supervised  logistic models. 
In Section 4, Monte Carlo simulations and benchmark data analysis are given to assess the performances of our proposed semi-supervised logistic discrimination. 
Some concluding remarks are given in Section 5.

\section{Semi-supervised logistic modeling from different sampling distributions}


\subsection{Linear logistic modeling for semi-supervised learning}
\label{Linear logistic modeling for semi-supervised learning}

We review here semi-supervised linear logistic models developed by early researchers (e.g., Amini and Gallinari, 2002; Vittaut {\it et al.}, 2002). 
Suppose that we have an $n_1$ labeled data set $\{ ({\bm x}_{\alpha}, y_{\alpha}); \alpha=1,\ldots,n_1 \}$ and an $(n - n_1)$ unlabeled data set $\{ {\bm x}_{\alpha}; \alpha=n_1+1,\ldots, n \}$, where ${\bm x}_{\alpha} = (x_{\alpha1}, \ldots, x_{\alpha p})^T$ denotes a $p$-dimensional explanatory variable and $Y_{\alpha}$ is a random variable taking values 0 or 1 with probabilities
\begin{eqnarray}
{\rm Pr} (Y_{\alpha} = 1 | {\bm x}_{\alpha}) = \pi ({\bm x}_{\alpha}), \qquad {\rm Pr} (Y_{\alpha} = 0 | {\bm x}_{\alpha}) = 1-\pi ({\bm x}_{\alpha}).
\label{posterior probability}
\end{eqnarray}
Note that logistic models are first constructed by only labeled data set, while the unlabeled data set is used in estimating the parameters involved in the logistic models.

Using conditional probabilities in Equation (\ref{posterior probability}) and the labeled data set, a linear logistic model (see, e.g., Hastie {\it et al.}, 2009) is formulated by
\begin{eqnarray}
\log \left\{ \frac{\pi ({\bm x}_{\alpha})}{1 - \pi ({\bm x}_{\alpha})} \right\} = w_{0} + \sum_{j=1}^p w_{j} {x}_{\alpha j} = {\bm w}^{T}  {\bm x}^*_{\alpha}, \quad \alpha=1,\ldots, n_1,
\label{linear logistic model}
\end{eqnarray}
where ${\bm w} = (w_{0}, w_{1}, \ldots,w_{p})^{T}$ is an unknown parameter vector and ${\bm x}_\alpha^* = (1, {\bm x}_\alpha^T)^T$. 
Hereafter, we denote conditional probabilities by $\pi ({\bm x}_{\alpha} ; {\bm w})$, since the conditional probabilities depend on the parameter vector ${\bm w}$. 
It follows from Equation (\ref{linear logistic model}) that conditional probabilities can be rewritten as
\begin{eqnarray}
\pi ({\bm x}_\alpha ; {\bm w}) = \frac{\exp ({\bm w}^T {\bm x}_\alpha^*)}{1 + \exp ({\bm w}^T {\bm x}_\alpha^*)}.
\label{posterior prob 2}
\end{eqnarray}
Also, a probability function of the random variable $Y_\alpha$ is the Bernoulli distribution in the form
\begin{eqnarray}
f (y_{\alpha} | {\bm x}_{\alpha}; {\bm w}) = \pi ({\bm x}_{\alpha} ; {\bm w})^{y_{\alpha}} \{ 1 - \pi ({\bm x}_{\alpha} ; {\bm w}) \}^{1 - y_{\alpha}}, \qquad y_{\alpha} = 0, 1.
\label{pdf1}
\end{eqnarray}
Under the linear logistic model, the log-likelihood function for $y_\alpha$ in terms of ${\bm w}$ is induced into
\begin{eqnarray}
\ell ({\bm w}) &=&\sum_{\alpha=1}^{n_1} \log f (y_{\alpha} | {\bm x}_{\alpha}; {\bm w}) \nonumber \\
&=& \sum_{\alpha=1}^{n_1} \left[ y_{\alpha} \log \pi ({\bm x}_{\alpha} ; {\bm w}) + (1-y_{\alpha}) \log \{ 1 - \pi ({\bm x}_{\alpha} ; {\bm w}) \} \right] \nonumber \\
&=&  \sum_{\alpha=1}^{n_1} \left[ y_{\alpha} {\bm w}^T {\bm x}_{\alpha}^* - \log \{ 1 + \exp ({\bm w}^T  {\bm x}_{\alpha}^*) \}\right].
\label{log-likelihood}
\end{eqnarray}

The unknown parameter $\bm w$ included in the logistic model is usually estimated by maximizing the log-likelihood function with respect to the parameter. 
The procedure is known as the supervised learning, i.e., the parameter is determined by using only labeled data set. 
Since we have an additional unlabeled data set, the parameter should be estimated by both labeled and unlabeled data set, which is called  the semi-supervised learning. 
Thereby, Amini and Gallinari (2002) proposed a log-likelihood function with additional unlabeled data given by
\begin{eqnarray}
\ell^* ({\bm w}) &=& \sum_{\alpha=1}^{n_1} \left[ y_{\alpha} {\bm w}^T {\bm x}_{\alpha}^* - \log \{ 1 + \exp ({\bm w}^T  {\bm x}_{\alpha}^*) \}\right] \nonumber \\
&& +  \sum_{\alpha=n_1+1}^{n} \left[ t_{\alpha} {\bm w}^T {\bm x}_{\alpha}^* - \log \{ 1 + \exp ({\bm w}^T  {\bm x}_{\alpha}^*) \}\right],
\label{linear semi-supervised}
\end{eqnarray}
where $t_{\alpha} \ (\alpha = n_1+1,\ldots, n)$ is a latent variable coded as 0 or 1. 
Amini and Gallinari (2002) estimated the parameter by maximizing the Equation (\ref{linear semi-supervised}) with the technique of EM algorithm, while Kawano and Konishi (2011) employed the Equation (\ref{linear semi-supervised}) with a regularization term in estimating the parameter in the context of nonlinear logistic models based on basis expansions.

Given the estimate $\hat{\bm w}$, we assign a future observation ${\bm x}_{\rm f}$ into class $j \ (j = 0, 1)$ that has the maximum conditional probability in the Equation (\ref{posterior prob 2}).

\subsection{Semi-supervised logistic model for different distributions}

Logistic models for semi-supervised learning described in Section \ref{Linear logistic modeling for semi-supervised learning} usually assumes that a density function for the labeled data set is the same as that for the unlabeled data set, i.e., when we denote that $q_{\rm label} ({\bm x})$ is a probability density function of explanatory variables for the labeled data and $q_{\rm unlabel} ({\bm x})$ is that for the unlabeled data, $q_{\rm label} ({\bm x}) = q_{\rm unlabel} ({\bm x})$. 
Our aim in this section is to construct logistic models under the situation that a density for the labeled data set is different from that for the unlabeled data set, i.e., $q_{\rm label} ({\bm x})  \neq q_{\rm unlabel} ({\bm x})$.

We recall the log-likelihood function for logistic models with unlabeled data in Equation (\ref{linear semi-supervised}). 
For the log-likelihood function, we propose a weighted log-likelihood function with unlabeled data in the form
\begin{eqnarray}
\ell^* ({\bm w}; \gamma_1, \gamma_2)&=&\sum_{\alpha=1}^{n_1} \left\{ \frac{q_{\rm unlabel} ({\bm x}_{\alpha})}{q_{\rm label} ({\bm x}_{\alpha})} \right\}^{\gamma_1} \left[ y_{\alpha}  {\bm w}^T {\bm x}_{\alpha}^* - \log \{ 1 + \exp ({\bm w}^T {\bm x}_{\alpha}^*) \} \right] \nonumber \\
&& \quad + \sum_{\alpha=n_1+1}^{n} \left\{ \frac{q_{\rm label} ({\bm x}_{\alpha})}{q_{\rm unlabel} ({\bm x}_{\alpha})} \right\}^{\gamma_2} \left[ t_{\alpha} {\bm w}^T {\bm x}_{\alpha}^* - \log \{ 1 + \exp ({\bm w}^T {\bm x}_{\alpha}^*) \} \right],
\label{semi cov lambda}
\end{eqnarray}
where $\gamma_1, \gamma_2 \in [0,1]$ are tuning parameters. 
If both $\gamma_1$ and $\gamma_2$ are 0, the log-likelihood function in Equation (\ref{semi cov lambda}) coincides with that in Equation (\ref{linear semi-supervised}). 
Note that the weight on the first term, $q_{\rm unlabel} ({\bm x}) / q_{\rm label} ({\bm x})$, is bigger near high densities of unlabeled data compared to those of labeled data, while that on the second term, $q_{\rm label} ({\bm x}) / q_{\rm unlabel} ({\bm x})$, is strengthen near high densities of labeled data compared to those of unlabeled data. 
Hence, the log-likelihood function on the first term is highly weighted near high densities of unlabeled data compared to those of labeled data, while that on the second term has high weighting near high densities of labeled data compared to those of unlabeled data. 
An idea of the weight, the ratio of $q_{\rm label} ({\bm x})$ and $q_{\rm unlabel} ({\bm x})$, arises from a statistical inference under covariate shift (Shimodaira, 2000). 
In the semi-supervised learning, employing a ratio of densities in log-likelihood functions is not new. 
For example, Kawakita and Kanamori (2012), Sokolovska {\it et al.} (2008), and Zou {\it et al.} (2007) use a ratio of densities in the semi-supervised inference. 
However, the Equation (\ref{semi cov lambda}) is a novel formulation in the semi-supervised context.

The Equation (\ref{semi cov lambda}) includes unknown values of ratios, $q_{\rm unlabel} ({\bm x}) / q_{\rm label} ({\bm x})$ and $q_{\rm label} ({\bm x}) / q_{\rm unlabel} ({\bm x})$, which are to be estimated. 
Various researchers address the problem of estimating the ratios by using several methods of statistics or machine learning (Bickel {\it et al.}, 2009; Huang {\it et al.}, 2007; Kanamori {\it et al.}, 2009; Sugiyama {\it et al.}, 2008; Sugiyama and Kawanabe, 2012; Sugiyama {\it et al.}, 2012). 
In this paper, we employ a uLSIF method proposed by Kanamori {\it et al.}  (2009) in determining values of the ratios, where the determination is performed before estimating the parameter $\bm w$. 
Also, a source code of  the method uLSIF is available in {\it http://www.math.cm.is.nagoya-u.ac.jp/\~{}kanamori/software/LSIF}. 
We do not follow details of the density ratio estimation procedure by the uLSIF method, since these are not our focus in this paper. 
For readers that are interested in the topics, we refer to Kanamori {\it et al.}  (2009), and Sugiyama and Kawanabe  (2012).

\subsection{Parameter estimation via regularization}
In estimating parameters in logistic models, the log-likelihood function often diverges to infinity when the maximum likelihood method is applied (Konishi and Kitagawa, 2008). 
Hence, the parameter vector $\bm w$ in Equation (\ref{semi cov lambda}) is estimated by the regularization method. 
The regularization method is to maximize a following regularized log-likelihood function
\begin{eqnarray}
\ell_\lambda^*({\bm w} ; \gamma_1, \gamma_2) = \ell^*({\bm w} ; \gamma_1, \gamma_2) - \frac{n_1 \lambda}{2} {\bm w}^T K {\bm w},
\label{regularized weighted likelihood}
\end{eqnarray}
where $\lambda$ is a regularization parameter that has positive values and $K = {\rm diag} (0, I_p)$ is a $(p+1) \times (p+1)$ matrix. 
Here, the matrix $I_p$ is a $p$-dimensional identity matrix.

It is not easy to optimize the parameter involved in Equation (\ref{regularized weighted likelihood}), since the latent variables $t_\alpha \ (\alpha=n_1+1,\ldots,n)$ are unobserved. 
Hence, we employ an EM-based algorithm developed by Kawano and Konishi (2011) as follows: 
\begin{description}
\item[Step1] Estimate the parameter vector ${\bm w}$  by maximizing the regularized log-likelihood function using only labeled data set $\{ ({\bm x}_{\alpha}, y_{\alpha}) ; \alpha = 1,\ldots,n_1 \}$ along with the technique of Newton-Raphson method. 
\item[Step2] Construct a classification rule $\pi ({\bm x}_{\alpha} ; \hat{\bm w})$. 
\item[Step3] (E-step) According to the classification rule in Step2, compute the conditional probabilities $\pi ({\bm x}_{\alpha} ; \hat{\bm w})$ for unlabeled data  ${\bm x}_{\alpha} \ (\alpha = n_1+1,\ldots,n)$. 
By using the conditional probabilities, estimate ${ t}_{\alpha}$ in the form  $\hat{t}_{\alpha} =\pi ({\bm x}_{\alpha} ; \hat{\bm w})$.
\item[Step4] (M-step) Replace $t_\alpha$ into $\hat{t}_\alpha$ in the regularized log-likelihood function (\ref{regularized weighted likelihood}), and then determine the parameter vector ${\bm w}$ through the maximization of the log-likelihood function in Equation (\ref{regularized weighted likelihood}) with the help of Newton-Raphson method. 
\item[Step5] Repeat the Step2 to the Step4 until the following condition
\begin{eqnarray}
| \ell_\lambda^* (\hat{\bm w}^{(k+1)} ; \gamma_1, \gamma_2) - \ell_\lambda^* (\hat{\bm w}^{(k)} ; \gamma_1, \gamma_2) | < \varepsilon
\end{eqnarray}
is satisfied, where $\hat{\bm w}^{(k)}$ is the value of ${\bm w}$ after the $k$-th EM iteration and $\varepsilon$ is an arbitrary small number (e.g., $10^{-5}$).  
\end{description}

It follows from these procedures that we obtain a statistical model in the form
\begin{eqnarray}
f(y | {\bm x} ; \hat{\bm w}) = \pi ({\bm x} ; \hat{\bm w})^y \{1-  \pi ({\bm x} ; \hat{\bm w}) \}^{1-y}.
\label{statistical model}
\end{eqnarray}
Note that the statistical model is constructed by using both labeled data and unlabeled data.

\section{Model selection criterion}
The statistical model in Equation (\ref{statistical model}) contains some adjusted parameters including two tuning parameters $\gamma_1, \gamma_2$ in the weighted log-likelihood function and the regularization parameter $\lambda$. 
Regarding the selection of these adjusted parameters as that of candidate models, we introduce a model selection criterion from an information-theoretic approach.

Let $y_1, \ldots, y_{n_1}$ be $n_1$ observations drawn randomly from an unknown probability distribution function $G(y|x)$ having a density function $g(y|x)$. 
On the other hand, we assume that $n_1$ observations for explanatory variables ${\bm x}_1, \ldots, {\bm x}_{n_1}$ are non-random; i.e., ${\bm x}_1, \ldots, {\bm x}_{n_1}$ are fixed (for details of this assumption, we refer to Konishi and Kitagawa, 2008). 
Under these settings, we derive a model selection criterion from the viewpoint of information theory.

Suppose that ${\bm z} = (z_1, \ldots, z_{n_1})^T$ are future observations for the response variable generated from $g(y|x)$. 
Let $f({\bm z} | {\bm x} ; \hat{\bm w}_G)^{\eta ({\bm x})} = \prod_{\alpha=1}^{n_1} f({z}_{\alpha} | {\bm x}_{\alpha} ; \hat{\bm w}_G)^{\eta ({\bm x}_{\alpha})}$  and $g({\bm z} | {\bm x}) = \prod_{\alpha=1}^{n_1} g({z}_{\alpha} | {\bm x}_{\alpha} )$, where  $\hat{\bm w}_G$ is an estimator of the parameter by any estimation procedures, $\eta ({\bm x}) = \eta ({\bm x}_1) + \cdots + \eta ({\bm x}_{n_1})$, and $\eta ({\bm x}_\alpha) \ (\alpha=1,\ldots,n_1)$ are weights that depend on explanatory variables ${\bm x}_\alpha$, which satisfy $\eta ({\bm x}_\alpha) > 0$. 
Note that the weights $\eta ({\bm x}_\alpha) \ (\alpha=1,\ldots,n_1)$ are fixed, since we assume that ${\bm x}_1, \ldots, {\bm x}_{n_1}$ are non-random. 
Then Irizarry (2001) implicitly proposes a following Kullback--Leibler information in order to measure the divergence of the statistical model with weights from the true distribution: 
\begin{eqnarray}
I \{ g ; f \} &=& E_{G({\bm z} | {\bm x})} \left[ \log \frac{g({\bm z} | {\bm x})}{f({\bm z} | {\bm x} ; \hat{\bm w}_G)^{\eta ({\bm x})} } \right] \nonumber \\
&=& E_{G({\bm z} | {\bm x})} \left[ \log g({\bm z} | {\bm x}) \right] - E_{G({\bm z} | {\bm x})} \left[ \log f({\bm z} | {\bm x} ; \hat{\bm w}_G)^{\eta ({\bm x})} \right] \nonumber \\ 
&=& E_{G({\bm z} | {\bm x})} \left[ \log g({\bm z} | {\bm x}) \right] - E_{G({\bm z} | {\bm x})} \left[ {\eta ({\bm x})} \log f({\bm z} | {\bm x} ; \hat{\bm w}_G) \right]. 
\label{KLdiv}
\end{eqnarray}
The best model can be regarded as the best minimizer of the Kullback--Leibler information (Irizarry, 2001). 
Since the first term of Equation (\ref{KLdiv}) does not depend on the models with the estimator $\hat{\bm w}_G$, we have only to consider the second term of Equation (\ref{KLdiv}). 
Therefore, we focus on maximizing the second term of Equation (\ref{KLdiv}) which leads to the minimization of the Kullback--Leibler information.

By introducing an estimator of the second term of Equation (\ref{KLdiv}),  a model selection criterion is, generally, given by 
\begin{eqnarray}
{\rm IC} = -2 \sum_{\alpha=1}^{n_1}  {\eta ({\bm x}_\alpha)} \log f({y}_\alpha | {\bm x}_\alpha ; \hat{\bm w}_G) + 2 \hat{b} (G),
\label{IC}
\end{eqnarray}
where IC stands for information criterion and $\hat{b}(G)$ is an estimator of the bias $b(G)$ in the following: 
\begin{eqnarray}
b (G) = E_{G({\bm y} | {\bm x})} \left[ { \sum_{\alpha=1}^{n_1} \eta ({\bm x}_\alpha)} \log f (y_{\alpha} | {\bm x}_{\alpha} ; \hat{\bm w}_G) - E_{G({\bm z} | {\bm x})} \left[ {\eta ({\bm x})} \log f ({\bm z} | {\bm x} ; \hat{\bm w}_G) \right] \right].
\label{bias}
\end{eqnarray}

Suppose that the estimator $\hat{\bm w}_M$ of the parameter is an M-estimator defined as the solution of the following implicit equation:
\begin{eqnarray}
\sum_{\alpha=1}^{n_1} {\bm \psi} (y_{\alpha} | {\bm x}_\alpha ; \hat{\bm w}_M) = {\bm 0}
\end{eqnarray}
with $\bm \psi$ being referred to as $\bm \psi$--function (e.g., see, Huber, 2004). 
Using the idea of Konishi and Kitagawa (1996), we  derive a model selection criterion for the statistical models with the M-estimator $\hat{\bm w}_M$ in the form
\begin{eqnarray}
{\rm IC}_M =  -2 \sum_{\alpha=1}^{n_1}  {\eta ({\bm x}_\alpha)} \log f({y}_\alpha | {\bm x}_\alpha ; \hat{\bm w}_M) + 2 {\rm tr} \left\{ Q(\hat{\bm w}_M) R^{-1} (\hat{\bm w}_M)\right\},
\end{eqnarray}
where $Q(\hat{\bm w}_M)$ and $R(\hat{\bm w}_M)$ are given by
\begin{eqnarray}
Q(\hat{\bm w}_M) &=& \frac{1}{n_1} \sum_{\alpha=1}^{n_1} {\bm \psi} (y_{\alpha} | {\bm x}_\alpha ; {\bm w}) \displaystyle{\frac{\eta ({\bm x}_{\alpha}) \partial  \log f(y_\alpha | {\bm x}_\alpha ; {\bm w}) }{\partial \bm w^T} \bigg|_{{\bm w} = \hat{\bm w}_M}}, \\\
R(\hat{\bm w}_M) &=& - \frac{1}{n_1} \sum_{\alpha=1}^{n_1} \displaystyle{  \frac{\partial {\bm \psi} (y_{\alpha} | {\bm x}_\alpha ; {\bm w})^T}{\partial {\bm w}}\bigg|_{{\bm w} = \hat{\bm w}_M}  }.
\end{eqnarray}

In our models, the estimator $\hat{\bm w}$, which maximizes the regularized log-likelihood function in Equation (\ref{regularized weighted likelihood}), can be regarded as an M-estimator. 
Here, we set the $\bm \psi$--function of the estimator into
\begin{eqnarray}
{\bm \psi} (y_{\alpha} | {\bm x}_\alpha ; {\bm w}) &=& \frac{\partial}{\partial {\bm w}} \left[ \left\{ \frac{q_{\rm unlabel} ({\bm x}_{\alpha})}{q_{\rm label} ({\bm x}_{\alpha})} \right\}^{\gamma_1} \left[ y_{\alpha}  {\bm w}^T {\bm x}_{\alpha}^* - \log \{ 1 + \exp ({\bm w}^T {\bm x}_{\alpha}^*) \} \right] - \frac{\lambda}{2}  {\bm w}^T K {\bm w} \right]. \nonumber \\
\label{psi-function}
\end{eqnarray}
Note that the $\bm \psi$--function in Equation (\ref{psi-function}) is actually incorrect since the estimator $\hat{\bm w}$ is obtained by maximizing the Equation (\ref{regularized weighted likelihood}) with respect to the parameter; i.e., the estimator are constructed by using both labeled and unlabeled data. 
However, $\bm \psi$--functions in the context of model selection criteria must be given by a regularized or non-regularized log-likelihood function with incomplete data; i.e., the functions does not include latent variables (for details, see, Hirose {\it et al.}, 2008). 
Hence, we employ the $\bm \psi$--function in Equation (\ref{psi-function}) in order to derive a model selection criterion.

By using the $\bm \psi$--function in Equation (\ref{psi-function}) and substituting $\{ q_{\rm unlabel} ({\bm x}_{\alpha}) / q_{\rm label} ({\bm x}_{\alpha}) \}^{\gamma_1}$ for the weights ${\eta} ({\bm x}_\alpha) \ (\alpha=1,\ldots,n_1)$, we introduce a generalized information criterion (GIC) for evaluating our proposed semi-supervised logistic models estimated by the regularization method. 
The model selection criterion is given by
\begin{eqnarray}
{\rm GIC} = -2 \sum_{\alpha=1}^{n_1} \left\{ \frac{q_{\rm unlabel} ({\bm x}_\alpha) }{q_{\rm label} ({\bm x}_\alpha) } \right\}^{\gamma_1} \log f (y_{\alpha} | {\bm x}_{\alpha} ; \hat{\bm w}) + 2 {\rm tr} \left\{ Q(\hat{\bm w}) R^{-1}(\hat{\bm w}) \right\},
\label{GIC}
\end{eqnarray}
where the matrices $Q (\hat{\bm w})$ and $R (\hat{\bm w})$ are
\begin{eqnarray}
Q (\hat{\bm w}) &=& \frac{1}{n_1} \left\{ X^T \hat{W}^2 \hat{\Lambda}^2 X -  \lambda K \hat{\bm w} {\bm 1}_{n_1}^T \hat{W} \hat{\Lambda} X \right\}, \\
R (\hat{\bm w}) &=& \frac{1}{n_1} X \hat{\Pi} \hat{W} ({ I}_{n_1} - \hat{\Pi}) X + \lambda K.
\end{eqnarray}
Here, $\bm 1_{n_1}$ is an $n_1$-dimensional vector of which the elements are all one, and $I_{n_1}$ is an $n_1$-dimensional identity matrix. 
Also, $X, \hat{W}, \hat{\Lambda}$, and $\hat{\Pi}$ are, respectively, given by
\begin{eqnarray*}
X &=& ({\bm x}_1^*, \ldots, {\bm x}_{n_1}^*)^T, \\
\hat{W} &=& {\rm diag} \left[ \left\{ \frac{q_{\rm unlabel} ({\bm x}_1) }{q_{\rm label} ({\bm x}_1) } \right\}^{\gamma_1}, \ldots ,\left\{ \frac{q_{\rm unlabel} ({\bm x}_{n_1}) }{q_{\rm label} ({\bm x}_{n_1}) } \right\}^{\gamma_1} \right], \\
\hat{\Lambda} &=& {\rm diag} \left[ y_1 - \pi({\bm x}_1 ; \hat{\bm w}), \ldots, y_{n_1} - \pi({\bm x}_{n_1} ; \hat{\bm w}) \right], \\
\hat{\Pi} &=& {\rm diag} \left[ \pi({\bm x}_1 ; \hat{\bm w}), \ldots,  \pi({\bm x}_{n_1} ; \hat{\bm w}) \right].
\end{eqnarray*}
Note that the GIC in Equation (\ref{GIC}) seemingly appears not to depend on all adjusted parameters (in particular, $\gamma_2$). 
However, the GIC implicitly includes the adjusted parameters $(\lambda, \gamma_1, \gamma_2)$, since the estimator $\hat{\bm w}$ depends on all adjusted parameters.

We choose the adjusted parameters from the minimizer of the GIC in Equation (\ref{GIC}).

\section{Numerical studies}

We studied some numerical examples to show the efficiency of our proposed modeling strategy. 
Two types of Monte Carlo simulations and benchmark data analysis are given to illustrate the proposed semi-supervised logistic discrimination. 

\subsection{Simulation 1}

We investigated the effectiveness of the proposed modeling procedures through Monte Carlo simulations. 
In this simulation study, we generated data sets $\{ ({x}_{1\alpha}, {x}_{2\alpha}, y_\alpha); \alpha=1,\ldots,n \}$ as labeled data and $\{ ( x_{1\alpha}, x_{2\alpha} ); \alpha=1,\ldots,500 \}$ as unlabeled data. 
In labeled data,  $( x_{1\alpha}, x_{2\alpha} )$ were generated by a normal distribution $N((-0.9,1-\sin(\sin( 0.9^2 \pi)))^T, {\rm diag} \\ (0.0015, 2))$, and  $y_\alpha$ was generated according to a following conditional probability
\begin{eqnarray}
{\rm Pr} (Y=1 |  x_{1}, x_{2} ) = 1/ \left[ 1 + \exp\left\{ -\sin (2\pi x_1^2) - x_2 + 1 \right\} \right].
\label{conditional probability}
\end{eqnarray}
Meanwhile, unlabeled data $( x_{1\alpha}, x_{2\alpha} )$ were obtained by a normal distribution $N((-0.4,1-\sin(\sin( 0.4^2 \pi)))^T, {\rm diag} (0.05, 1))$.
Test data  $\{ ({x}_{1\alpha}, {x}_{2\alpha}, y_\alpha); \alpha=1,\ldots,1000 \}$ were generated as follows. 
First, $ ({x}_{1\alpha}, {x}_{2\alpha})$ were derived by a mixture of labeled and unlabeled data, where the mixing rate is equal (that is, 0.5). 
Second, for the $ ({x}_{1\alpha}, {x}_{2\alpha})$, $y_\alpha$ was obtained according to the conditional probability in Equation (\ref{conditional probability}). 
We assumed that labeled data sizes ($n$) were 25, 50, 100, 150, 200, and 250.

We fitted our semi-supervised logistic regression model to the data sets. 
Note that the density ratio estimation procedure by the uLSIF method described in Section 2.2 is not performed in these simulation trials, since the density ratio is exactly calculated. 
The simulation results were obtained by averaging over 50 repeated Monte Carlo trials. 
For each data set, we computed averages of prediction error rates (PE) for 50 iterations. 
The tuning parameters in our models were selected by using the GIC in Equation (\ref{GIC}). 
For 50 trials, we computed averages of selected adjusted parameters. 
The results are summarized in Table \ref{Simulation1}.  
From the table, in the selection of adjusted parameters, the values of the tuning parameter $\gamma_1$ are 0.10 in all cases, while those of the parameter $\gamma_2$ increase with the increasing numbers of labeled data.  
The regularization parameter $\lambda$ takes smaller values according to the increasing numbers of labeled data. 

\begin{table}[t]
\begin{center}
\caption{Comparison of prediction error rates ($\%$) and values of selected parameters for several number of labeled data points. }
\vspace{5mm}
\begin{tabular}{lcccccccc}
\hline
Method $\setminus$ \# of labeled data & & 25 & 50 & 100 & 150 & 200 & 250 \\ \hline
SSLRCS        & PE & 33.3 & 33.3  & 33.9 &  34.8 &  35.5 &  35.0  \\
                        & $\log_{10}(\lambda)$ & --2.20 & --3.00  & --3.18 &  --3.54 &  --3.80 &  --3.72  \\
                        & $\gamma_1$ & 0.10 & 0.10  & 0.10 &  0.10 &  0.10 &  0.10  \\
                        & $\gamma_2$ & 0.61 & 0.71  & 0.74 &  0.82 &  0.86 &  0.82  \\
LSSLR   &PE & 34.3 & 34.4 &  34.2 &  35.3 &  35.9 &  35.6  \\
                        & $\log_{10}(\xi_1)$ & --2.72 & --3.36  & --3.38 &  --3.72 &  --3.88 &  --3.92  \\
SLR      &PE & 35.6 &  34.3 &  34.3 &  35.2 &  35.8 &  35.6   \\
                        & $\log_{10}(\xi_2)$ & --2.06 & --2.32  & --2.80 &  --3.10 &  --3.50 &  --3.68  \\
\hline
\end{tabular}
\label{Simulation1}
\end{center}
\end{table}

We compared the performances of the proposed semi-supervised methodologies (SSLRCS: semi-supervised logistic regression under covariate shift) with those of semi-supervised method proposed by  Amini and Gallinari (2002) (LSSLR: linear semi-supervised logistic regression), which is developed under the condition that density functions for labeled and unlabeled data are same, and supervised linear logistic discriminant analysis (SLR: supervised logistic regression). 
Note that the SLR is constructed by using  only labeled data. 
Semi-supervised and supervised logistic modeling strategies were applied into the data sets. 
The LSSLR and the SLR include a tuning parameter, respectively, where we denote the tuning parameters as $\xi_1$ and $\xi_2$, respectively. 
The parameter is determined by the GIC, where the GIC for LSSLR is obtained by setting $q_{\rm unlabel} ({\bm x}_\alpha)/q_{\rm label} ({\bm x}_\alpha) = 1\ (\alpha=1,\ldots,n_1)$ in Equation (\ref{GIC}) and that for SLR is given by Ando {\it et al.} (2008). 
For these methods, we also computed averages of prediction error rates and selected tuning parameters. 
It may be seen from Table \ref{Simulation1} that SSLRCS is superior to other methods (LSSLR and SLR) in all cases in the sense that the proposed method gives smaller prediction error rates.

\subsection{Simulation 2}

We simulated three data sets given in Chakraborty (2011) to examine the performances of our proposed modeling strategy. 
For each of the simulation cases, we generated 100 data points in the labeled data set, 1000 data points in the unlabeled data set, and 1000 data points in the test data set. 
Using the data sets, we constructed the SSLRCS, the LSSLR, and the SLR. 
We repeated the procedure 50 times. 
Our simulation settings are given as follows (for details, see, Chakraborty (2011, p. 76)):
\begin{itemize}
\item Case 1 : In the labeled data set, generate ${\bm x} = (x_1, x_2)^T$ given by $x_i \sim N(2,1) \ (i=1,2)$ for Class 1 and $x_i \sim N(-2,1) \ (i=1,2)$ for Class 2. 
In the unlabeled data set, $x_i \sim N(2,2) \ (i=1,2)$ for Class 1 and $x_i \sim N(-2,2) \ (i=1,2)$ for Class 2. 
In the test data set, $x_i \sim 0.5N(2,1) + 0.5N(2,2) \ (i=1,2)$ for Class 1 and $x_i \sim 0.5N(-2,1) + 0.5N(-2,2) \ (i=1,2)$ for Class 2. 
\item Case 2 : Generate ${\bm x} = (x_1, \ldots, x_{10})^T$ given by $x_i \sim N(1,3) \ (i=1,\ldots,10)$ for Class 1 and $x_i \sim N(-1,3) \ (i=1,\ldots,10)$ for Class 2. 
\item Case 3 : Generate ${\bm x} = (x_1, x_2)^T$ given by $x_i \sim N(5,2) \ (i=1,2)$ for Class 1 and $x_i \sim N(8,2) \ (i=1,2)$ for Class 2 in the labeled data set. 
In the unlabeled data set, $x_i \sim N(6,2) \ (i=1,2)$ for Class 1 and $x_i \sim N(9,2) \ (i=1,2)$ for Class 2. 
In the test data set, $x_i \sim 0.5N(5,2) + 0.5N(6,2) \ (i=1,2)$ for Class 1 and $x_i \sim 0.5N(8,2) + 0.5N(9,2) \ (i=1,2)$ for Class 2. 
\end{itemize}

\begin{table}[t]
\begin{center}
\caption{Comparison of prediction error rates ($\%$) and values of selected parameters for several cases. }
\vspace{5mm}
\begin{tabular}{lcccc}
\hline
Method $\setminus$ Data sets && Case 1 & Case 2 & Case 3  \\ \hline
SSLRCS         &PE & 1.28  & 3.65  & 9.72   \\
                         &$\log_{10}(\lambda)$ & --2.50  & --1.98  & --1.98   \\
                         &$\gamma_1$ & 1.00  & 1.00  & 1.00   \\
                         &$\gamma_2$ & 0.102  & 0.106  & 0.106   \\
LSSLR   &PE& 1.36 & 4.19 &  11.6   \\
                &$\log_{10}(\xi_1$)& --2.50 & --2.00 &  --3.00   \\
SLR      &PE&  1.43 &  5.05 &  11.7    \\
                &$\log_{10}(\xi_2$)& --2.50 & --1.96 &  --2.18   \\
\hline
\end{tabular}
\label{Simulation2}
\end{center}
\end{table}

The results from the simulation studies are  in Table \ref{Simulation2}. 
We obtained the values in the table by averaging over 50 trials. 
The optimal tuning parameters selected by our model selection criterion were 1.00 for $\gamma_1$ in all situations, 0.102 and 0.106 for $\gamma_2$ in Case 1 and Case 2, 3, respectively, and $10^{-2.50}$ and $10^{-1.98}$ for $\lambda$ Case 1 and Case 2, 3, respectively. 
From the simulation results, we observe that our proposed procedure performs well in all cases with respect to minimizing prediction error rates even though Case 2 is an ordinary setting of semi-supervised learning, i.e., the density function for labeled data is same as that for unlabeled data. 
Hence, we conclude that our proposed method may be useful even if the densities for labeled and unlabeled data are same.

\subsection{Benchmark data analysis}

Thorough analyzing the g10 data set (Chapelle and Zien, 2005), the ionosphere data set (Sigillito {\it et al.}, 1989), and the pima data set (Ripley, 1996), we illustrated the effectiveness of the proposed semi-supervised methodology. 
The g10 data set includes 550 data points with 10 predictors, and we prepared 250 training data points and 300 test data points. 
The ionosphere data set consists of 356 data points with 33 predictors, and we split the whole 356 data points into 150 training data points and 206 test data points. 
The pima data set, which consists of 300 training data points and 232 test data points, is a binary classification with 7 predictors. 
In order to implement the semi-supervised procedure, the training data points were randomly split into two halves with labeled data points and unlabeled data points, where labeled data points were assigned as 5\%, 10\%, 20\%, 30\%, 40\%, and 50\% for training data points, respectively. 
We repeated the random splitting 50 times. 
We also compared our proposed method (SSLRCS) with the LSSLR and the SLR, which are described in Section 4.1.

Table \ref{benchmark} shows the summary of the prediction errors and selected adjusted parameters for the benchmark data sets. 
The values in the table were  averaged over 50 repetitions. 
From the results, we observe that the tuning parameter $\gamma_1$ provides the largest values (i.e., 1.00) in almost all cases, while the parameter $\gamma_2$ gives relatively smaller values (i.e., from 0.10 to 0.40). 
We also find that our proposed procedure outperforms the previously proposed methods in almost all situations, although it is unclear that whether densities for labeled and unlabeled data are different. 
In particular, the proposed method seems to work well when the number of labeled data points is small. 

\begin{table}[htbp]
\begin{center}
\caption{Comparison of prediction error rates ($\%$) and values of selected parameters for some data sets. }
\vspace{5mm}
\begin{tabular}{lcccccccc}
\hline
Method $\setminus$ \% && 5 & 10 & 20 & 30 & 40 & 50 \\ \hline
g10 \\
SSLRCS         &PE& 3.40 & 3.47  & 3.85 &  4.06 &  4.66 &  5.42  \\
                         &$\log_{10}(\lambda)$& --3.20 & --2.97  & --2.99 &  --3.00 &  --3.00 &  --3.00  \\
                         &$\gamma_1$& 1.00 & 1.00  & 1.00 &  1.00 &  1.00 &  1.00  \\
                         &$\gamma_2$& 0.15 & 0.10  & 0.10 &  0.10 &  0.10 &  0.10  \\
LSSLR  &PE & 26.6 & 16.2 &  9.94 &  7.04 &  5.66 &  4.77  \\
                         &$\log_{10}(\xi_1)$& --3.50 & --3.00  & --3.00 &  --3.00 &  --3.00 &  --3.00  \\
SLR     &PE & 26.4 &  16.4 &  9.30 &  6.85 &  5.45 &  4.62   \\
                         &$\log_{10}(\xi_2)$& --3.50 & --3.00  & --3.00 &  --3.00 &  --3.00 &  --3.00  \\
\hline
Ionosphere \\
SSLRCS         &PE& 18.2 & 17.3  & 16.9 &  16.4 &  17.3 &  16.8  \\
                         &$\log_{10}(\lambda)$& --2.89 & --2.86  & --2.70 &  --2.44 &  --2.61 &  --2.66  \\
                         &$\gamma_1$& 0.99 & 0.99  & 1.00 &  1.00 &  1.00 &  1.00  \\
                         &$\gamma_2$& 0.50 & 0.46  & 0.37 &  0.27 &  0.37 &  0.35  \\
LSSLR   &PE& 29.0 & 22.8 &  18.9 &  17.4 &  16.2 &  15.4  \\
                         &$\log_{10}(\xi_1)$& --3.92 & --3.50  & --3.50 &  --3.00 &  --3.00 &  --3.00  \\
SLR      &PE& 28.9 &  23.1 &  19.5 &  18.0 &  16.7 &  15.7   \\
                         &$\log_{10}(\xi_2)$& --3.92 & --3.50  & --3.50 &  --3.00 &  --3.00 &  --3.00  \\
\hline
Pima \\
SSLRCS         &PE& 26.6 & 26.9  & 26.6 &  26.8 &  26.7 &  26.7  \\
                         &$\log_{10}(\lambda)$& 1.41 & 1.53  & 1.35 &  1.30 &  1.27 &  1.36  \\
                         &$\gamma_1$& 1.00 & 1.00  & 1.00 &  1.00 &  1.00 &  1.00  \\
                         &$\gamma_2$& 0.30 & 0.28  & 0.26 &  0.23 &  0.24 &  0.23  \\
LSSLR   &PE& 30.1 & 27.0 &  27.0 &  27.0 &  26.9 &  26.7  \\
                         &$\log_{10}(\xi_1)$& 1.27 & 1.41  & 1.53 &  1.72 &  1.71 &  1.61  \\
SLR      &PE& 29.3 &  26.9 &  26.9 &  27.0 &  26.8 &  26.7   \\
                         &$\log_{10}(\xi_2)$& 2.46 & 2.37  & 2.34 &  2.23 &  2.16 &  2.10  \\
\hline
\end{tabular}
\label{benchmark}
\end{center}
\end{table}

\section{Concluding remarks}

We proposed a semi-supervised logistic classification methodology for different density functions of labeled and unlabeled data along with the technique of covariate shift adaptation and regularization. 
A crucial point for our semi-supervised modeling processes includes the choices of  some tuning parameters in our proposed models. 
We introduced a model selection criterion from the viewpoint of information theory in order to select the values of the adjusted parameters. 
Through Monte Carlo simulations and the benchmark data analysis, we showed that our modeling strategy is effectiveness in practical situations in  the viewpoints of yielding relatively lower prediction errors than previously developed methods. 
Our modeling procedure may be applied into the problem of constructing a nonlinear semi-supervised classification method based on basis expansions, which will be discussed in another paper. 

\vspace{5mm}
\noindent \textbf{\large Acknowledgement}}\\
This work was supported by the Ministry of Education, Science, Sports and Culture, Grant-in-Aid for Young Scientists (B), $\#$24700280, 2012--2015. 




\end{document}